\documentclass[a4paper]{sbgames}               
\usepackage{times}
\usepackage{graphicx}

\usepackage{parskip}

\usepackage[labelfont=bf,textfont=it]{caption}

\usepackage{url}

\title{On the Development of Intelligent Agents for MOBA Games}

\author{Victor do Nascimento Silva and Luiz Chaimowicz\\ \\Department of Computer Science\\Universidade Federal de Minas Gerais - Brazil}
\contactinfo{\{vnsilva,chaimo\}@dcc.ufmg.br}


\begin{document}

    \teaser{\includegraphics[width=\linewidth]{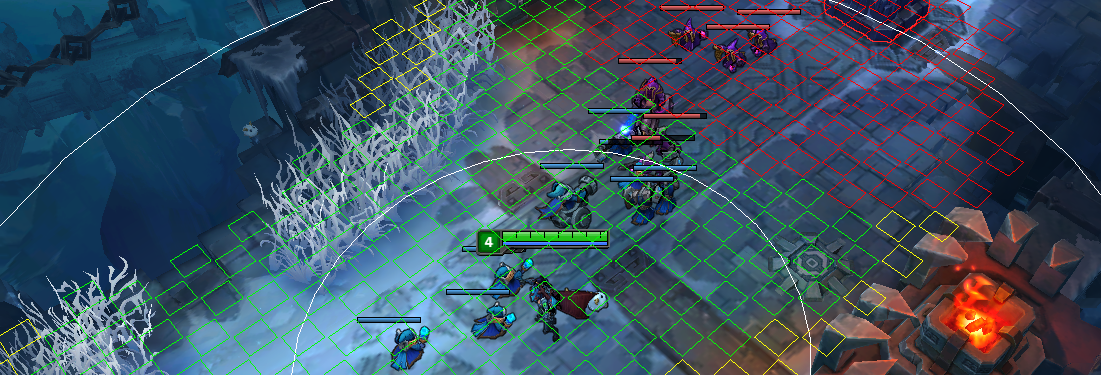}
}


\maketitle


\begin{abstract}
Multiplayer Online Battle Arena (MOBA) is one of the most played game genres nowadays. With the increasing growth of this genre, it becomes necessary to develop effective intelligent agents to play alongside or against human players. In this paper we address the problem of agent development for MOBA games. We implement a two-layered architecture agent that handles both navigation and game mechanics. This architecture relies on the use of Influence Maps, a widely used approach for tactical analysis. Several experiments were performed using {\em League of Legends} as a testbed, and show promising results in this highly dynamic real-time context.

\end{abstract}

\keywordlist
\contactlist

\section{Introduction}

The development of tactical movement systems is still a challenging task in game AI. This is especially true when dealing with competitive or combat scenarios. Behaving in a logical way in these cases could be very hard and require heavy computation and domain knowledge. Thus, it is necessary to collect data and deal with uncertainty when controlling agents in these scenarios to be able to behave in a competitive way. 

Researching tactical analysis can benefit not just games but areas like military analysis, robotics and traffic. Moreover, any area with partial information and dynamic environment can implement the techniques of tactical analysis towards improving agent behavior.

In game research, the problem of navigating in a combat scenario can be tackled by the development of specific strategies and tactical analysis. Furthermore, the tactical analysis field has gained much attention, as games become more complex. Tactical analysis was already performed in some agents, like Starcraft agents and ORTS competitions \cite{hagelback2008a},\cite{uriarte2012}. 

Another point that must be discussed is that MOBA is one of most played games of actuality, having almost 30\% of online computing gameplay time \footnote{http://bit.ly/1OopRvh}. However, there is not much attention from researchers over this game genre. Characteristics like the continuous updates can be cited as one of the causes of the low research rate over this domain. We aim to encourage researchers to use MOBA as a domain as testbed for their future research, providing gathered knowledge about this domain.

Games are a good option to serve as testbed for the development of AI. Having a limited environment that can reproduce almost all characteristics of the real world helps the research to perform reliable tests \cite{adobbati2001}. Games also provides an environment where winning, losing or drawing have well-defined rules, making it easy for AI to evaluate these criteria. Developing and testing AI algorithms in a game environment can benefit research by reducing the costs of test and real world resources.

Selecting a platform for testbed in AI research can be a hard task, especially when considering that most MOBA games are commercial. We considered several games in our initial research phase and ended with two main options: \textsc{Heroes of Newerth} (HoN) and \textsc{League of Legends}\footnote{http://www.riotgames.com/our-games} (LoL). When choosing between these two platforms we considered: (a) ease of implementation, in terms of language and documentation; (b) game community, in terms of number of players. Given these characteristics, we chose to develop our research using \textsc{League of Legends} as testbed.

In addition, real-time games are gaining much attention in recent years. We observe the growing of intelligent agents development for these games, like Starcraft. Moreover, there have been competitions using such agents, like the AIIDE and the CIG Starcraft agents competition \cite{ontanon2013}. However, in contrast to the popularity of MOBA, there is not a great attention related to this genre.

In this paper we address the problem of tactical analysis and navigation in combat scenarios. For doing so, we use a widely popular MOBA game, League of Legends, as testbed. Our approach consists in the development of an Intelligent Agent that aims to behave competitively in the chosen domain. Further, we implement a multi-layered architecture agent using Influence Maps (IM) and real-time reasoning. In addition we collect and provide information about the MOBA domain and we encourage researchers to adopt MOBA as their research domain.

In short, our contributions in this work are:
\begin{itemize}
\item We present MOBA games as testbed for AI;
\item Implementation of an game agent capable of winning the game;
\item An architecture for the game agent based in tactical analysis;
\item Our agent was capable of emerging the kiting behaviour without a dedicated module.
\end{itemize}

The rest of this paper is organized as follows: Section 2 introduces basic concepts about Tactical Analysis techniques, MOBA domain and LoL. In this section we also discuss related work in agent development and MOBA. Section 3 presents the architecture of the agent implemented, showing its characteristics and layers. Experiments performed for evaluating our approach is presented in Section 4. Lastly, Section 5 brings the conclusion and the directions for future work.


\section{Background }


Research in games requires collection of information about the domain to be used as guiding tools for the research and experiments to be developed. A game is composed of a set of rules during its execution and winning or losing criteria. Thus, understanding these characteristics is essential for implementation of agents, data and results collection. In this section we present substantial knowledge about techniques that were used in our research. Furthermore, we present the characteristics of the MOBA domain.

\subsection{Influence Maps and Potential Fields}

Tactical analysis has been studied by various researchers over time and was applied in a wide range of games in both academia and industry. The most common technique is to divide the game map into chunks and try to perform some kind of feature analysis in these chunks \cite{millington2012}. Analyzing the environment statically and dynamically in a consistent fashion can help the AI to behave in a smarter way. Also, the problem of coordinating agents in a tactical way is not new. From the work of \cite{reynolds1987}\cite{reynolds1999a}, we can observe the necessity of environment analysis when driving agents. Reynolds agents take into account the environment and other agents when driving themselves in a flock. Potential Fields (PF), a technique originally from Robotics area, tries to do such analyses, as presented by \cite{khatib1986} in his approach for real-time obstacle avoidance and agent driving. 

In games, Influence Maps (IM) is one the main methods applied for tactical analysis of the features in a map \cite{millington2012}. A general definition  of influence maps is presented in \cite{tozour2001}, which discusses its main concepts and advantages. Basically, the environment is decomposed in a grid, and each feature inserts points of influence in the cells where they are positioned. This influence is then spread using a proper algorithm, like Gaussian blur or Occlusion algorithm. Few years later, \cite{hagelback2008a} presented the use of Potential Fields in games, showing a similar approach as seen in Influence Maps. But, in spite of having the same name of the technique used in robotics \cite{khatib1986}, the work of \cite{hagelback2008a} divides the map in a grid, placing values over this grid that can be attractive or repulsive. 

The main difference between the approaches of IM and PF is that the first uses {\em spread} and {\em momentum} variables. These values can be used to modify the changes in information over time: while spread controls how far the value can reach, momentum controls the amount of time through which the information continues on the map. The use of such values allows IM to store historical information. Moreover, it turns IM into a versatile technique, as it can be used for both prediction and/or historical information just by performing parameter tuning \cite{champandard2011}. On other hand, PF turns to be very static, as its only parameter is the equation that controls the influence over the maps. Thus, we observe that PF is mainly used in real-time environments, especially for navigation.

\subsection{MOBA Games}

In recent years, the popularity of Real-Time Strategy has grown substantially. With them, a community of players interested in producing new content for such games has been created. These community is commonly called {\em Modders}, after the acronym of modifiers, generating the alternative game modes called modifications, or "mods". Among these, there was Aeon of Strife (AoS), the first MOBA like game, created as a Starcraft mod. 

After the popularity of AoS, modders used the Warcraft III engine to create a similar game mode and name it Defense of the Ancients (DotA). The DotA maps became popular among players, and there were many versions around. Its mechanics were broadly known that every other game that had similar gameplay was called "DotA like game" for a long time. The term MOBA was first observed in 2009 with the debuting title League of Legends, used by Riot Games to describe their game. Furthermore, MOBA games evolved from the LAN games to Multiplayer Online games, and even becoming more popular than MMORPGs, like World of Warcraft \cite{diamaranan2015}. Nowadays, MOBA is among the most played game genres in the world.

In terms of academic research, there is a rising interest in developing intelligent agents that are capable of playing games that run in real-time; have a dynamic environment; and provide partial information. The similarity of this kind of games to real-world problems is probably one of the main reasons for such research, as the solutions to these problems may be applied to various real-world scenarios \cite{tavares2014}. This growing interest for game agents, in particular of RTS games, allowed the academia to develop a novel game research scenario, creating competitions of intelligent game agents \cite{ontanon2013}.

Overall, it is very common to find two types of agents being studied: the perfect agent aims to defeat all other agents, and is the most common type. On the other hand, the human-like agents try to behave as humans in the game environment. This kind of agent is especially popular a in First Person Shooter (FPS) scenario, such as the \textsf{2kbotprize} competition~\cite{cothran2009}.

In spite of the agent developed in this work being a MOBA agent, we also discuss the development of agents for RTS, since, as far as we know, there were only a few works that try to develop an effective agent for MOBA. The work presented by \cite{weber2011} develops an human-behaving agent for \textsf{Starcraft}. In this work, the authors also discuss the development layers of an agent, classifying them in heterogeneous and homogeneous. An agent that is developed in just one layer is classified as homogeneous. This single layer implements all features of the agent, like navigation and actuation over the game. The work of \cite{uriarte2012} revisits the multi-layered system, implementing an agent that performs a kiting system using Influence Maps (IM). Moreover, the authors present a novel approach for tactical analysis, that was briefly discussed in \cite{hagelback2008b}.

In general tactical analysis, we find the work of \cite{stanescu2013}, in which the combat outcome is discussed in \textsf{Starcraft}. In the field of knowledge extraction and analysis, there is the work of \cite{bangay2014}, where the attributes are analysed in a RTS game and a framework for balance analysis is proposed. Furthermore, we believe that this knowledge extracted can be used to model tactical analysis in a combat scenario.

The MOBA research itself is a new field, with few published works. One of the reasons for this may be the lack of support in MOBA games to AI development. Most of the works found present general discussions or use collected data to extract information. The works of \cite{nosrati2013} and \cite{rioult2014}, for example, present a brief analysis of MOBA games, limited to features and specific characteristics. In the work of \cite{ferrari2013} we find a deep analysis of League of Legends, discussed by various aspects like Game Design, e-Sport and basic MOBA characteristics.

When analysing the gameplay and player abilities, is possible to find the work of \cite{drachen2014}, where the spatio-temporal skill of players is analyzed, focused in data visualization and gameplay analysis, using \textsf{Dota2} data. In the same research field, there is the work of \cite{yang2014} where, using \textsf{Dota2} matches log, the team-fights are analysed and used to learn patterns. These battles are modelled as interactivity graphs and later shaped in decision trees form. The model created is then used to predict the success of future battle in other matches. The work of \cite{pobiedina2013} performs a quantitative analysis of logs collected from \textsf{Dota2}, analyzing the relation between team formation and victories. It also performs an analysis of involved players, verifying that their experience and behavior during a match are essential to winning in MOBA.

Finally, the work of \cite{willich2015}, from our knowledge, is the only published work that addresses the problem of creating an agent for MOBA. The author uses Reinforcement Learning in HoN domain to make the agent learn how to behave in the game environment. The author shows that the developed agent learns the very basic mechanics of MOBA, and considers that further research is needed.

\subsection{MOBA agents}

For knowledge extraction from MOBA games, we choose to separate and simplify the game features towards a general model that can be used to achieve victory. First, we analyze the game model and steps that can allow an agent to win. Further, we analyze each feature in the agent's way to victory, dealing with each of them separately in a knowledge model, and then unifying them in an analytical model, in our case the Influence Map. 

MOBAs, as RTS, have a highly dynamic multi-agent environment. Instead of huge armies, MOBA focuses more in teamwork and strong micromanagement skills. The collaboration of multiple players pushes the team towards victory, although individual gameplay is also very valuable.

The characteristics of MOBA, in terms of gameplay, are also very similar to RTS. Listing these we can cite:

\begin{itemize}
\item Deterministic ambient: where the same actions have the same effect;
\item Partial information: there is a fog-of-war that covers areas out of agent sight line;
\item Non-persistent world: In contrast to most MMOGs, in MOBA every match the world is created; it does not exists when the match is over;
\item Dynamic/Real-time: The game world changes over time and accordingly to agent actions.
\end{itemize}

In addition, MOBA games goal consists in pushing a lane and destroying an enemy base. By analyzing various MOBAs, we can observe that very high-level sequence of three steps can be followed toward winning:

\begin{itemize}
\item Take enemy out of lane;
\item Push the lane;
\item Destroy structure;
\end{itemize}

Despite these high-level steps, its necessary a deep analysis to perform each of them. The lane pushing, for instance, requires tactical analysis of creeps, risk analysis, among others. Taking enemy out of lane requires tactical enemy analysis, micromanagement, among others. Structure destroying, on the other hand, requires enemy analysis, damage analysis and others. Furthermore, all the analysis must be performed real-time and with partial information. We discuss then how this analysis can be done and combined, generating a final knowledge tactical model.

We divided our analysis in three basic classes: Dynamic features, semi-dynamic features and static features. The dynamic features are the ones that can move through the scenario and perform actions. Semi-dynamic ones are units that, although do not move, can have a dynamic map values based in their states. Lastly, the static features are the ones that does not change through the game execution. We will discuss these classes further.

Dynamic units are the ones that can move through the scenario performing changes in the tactical analysis very fast. These units also can perform actions, like attacking or casting spells. The type of units that is contained in this class are: heroes, neutral creeps, enemy and allied creeps. In MOBA context, {\em creep} is a unit type that heroes can interact with but that they cannot control directly. This type of unit is AI-driven and is usually spawned in the team bases. Besides, neutral creeps are born in the jungle area, and do not move or attack unless it is injured by a hero unit. In some MOBAs, creeps can be also known as troops or minions. 

Heroes can be considered as the critical unit in dynamic class, they can change the game scenario very quickly, as most of their spells are casted instantly and they can be considered the fastest units in game. In addition, these units are responsible for the team fights and duels. Team fights are combats performed in groups of a team against another team, duels are fights that occur between two distinct players, one of each team.

Allied and enemy creeps are born in the respective team base. These units go into a lane and battle enemy creeps and structures pushing the lane towards enemy base. In some MOBAs, like Dota2, it is possible to both farm (kill enemy creeps) and deny (kill ally creeps). These units normally come in groups and have strong effect in tactical analysis, as they are very important in the initial phase of game, as heroes must perform last hits to obtain gold from them. Moreover, we found that it is necessary to verify the creeps that are most likely to be killed first, maximizing the effects of resource collection of players.

The only semi-dynamic unit contained in MOBA is the tower. The main characteristics of the semi-dynamic class is that it can modify its influence values based in its current aggression state. When attacking the controlled agent it generates negative values, as the agent should avoid receiving damage. On the other hand, the agent should attack the enemy towers, thus the influence generated is positive when there is another target being attacked by the tower.

Lastly, the static features of a MOBA game are considered as the terrain. The terrain cannot be modified over time, thus we can make an influence map that is static, performing the analysis just one time. Note that when performing static analysis we do not consider abilities of heroes that can cast some kind of collider or terrain modifier, considering these as dynamic characteristics.


\subsection{League of Legends}

\begin{figure}
\centering
\includegraphics[width=0.4\textwidth]{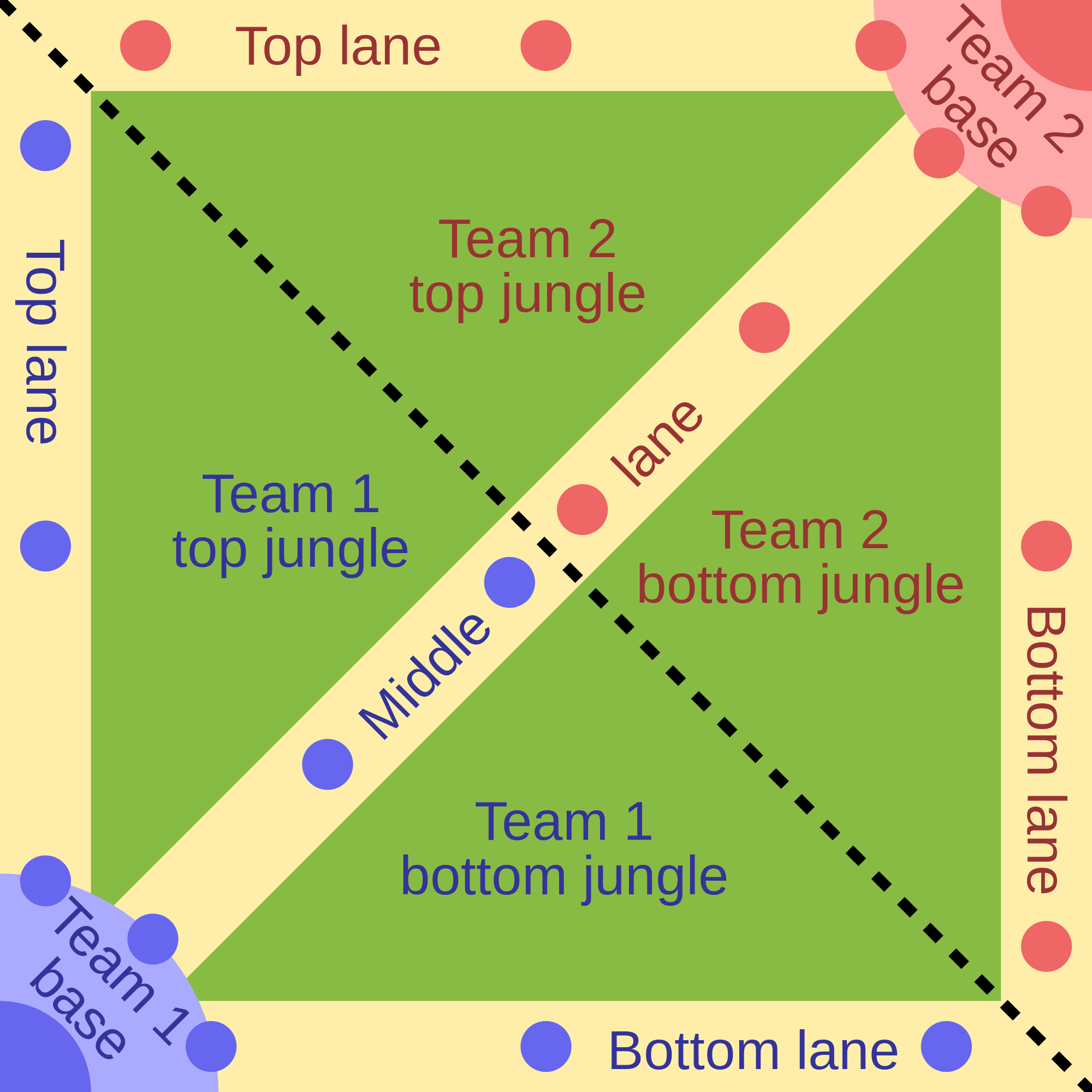}
\caption{General Map of a MOBA Game}
\label{img:mapofmoba}
\end{figure}

League of Legends is a game of Multiplayer Online Battle Arena (MOBA) genre developed by Riot Games; it was open released in 2009. MOBAs are characterized by its online competitive nature, where two teams of players compete against each other aiming to destroy the enemy base. A sample map can be seen in Figure \ref{img:mapofmoba}. In the most common game mode in MOBAs, each team is composed of five players. The player is represented by a powerful unit called {\em hero}, that is chosen at the beginning of each match from a hero pool. Each hero has unique powerful skills that can be used to combat enemy players, neutral units and creeps. Each hero starts at level one and the player has to defeat enemy, AI driven, weak units called creeps or jungle neutral units. By defeating these units, or enemy heroes, the player can accumulate experience points developing skills and upgrading their hero level. At each level, the player can upgrade one skill, making it more powerful, and the hero increases its power in terms of status, e.g. attack power, health points, mana points. When the player performs the last-hit in a neutral or enemy creep, his/her hero gets gold, allowing the player to buy items that will increase hero status, making it more powerful. Winning in a MOBA game can be achieved just by destroying an enemy main structure, in League of Legends this structure is called {\em Nexus}. When a team destroys the enemy Nexus the game is terminated and this is the winner team.

When choosing a hero, players must have in mind what role he/she wants to assume during gameplay. There are heroes that can be suited in various functions inside the game, but most of them performs better when assuming the main function that the hero was designed for. Another concept that the player must take in account is the current {\em metagame}. The metagame takes in account various theories about what the best compilation of heroes that tends to perform better in the actual scenario, using knowledge from inside and outside of the game. In addition, if the player is a human, his/her skills must be considered. Due to the continued development and update of MOBAs, the metagame tends to change in every balance change, done by a patch. An example of metagame is the most common team compilation in LoL, normally two players go to bottom lane, one Support and one Attack Damage Carry (ADC), in the middle lane goes a Mage or Assassin, a Tank or Fighter in the Jungle/Ganker role and a top lane player that can be an Initiator, Front-liner or Tank.

Metagame can be better understood by knowing the best functions of each class. An ADC is a long ranged hero, capable of dealing high damage to single targets, while a Support is a hero that can assist the ADC, healing of buffing his skills. Mages are champions that deal damage based in spells, and eventually are capable of performing high AoE damage. On the other hand, Assassins are strong in duels and killing a single enemy fast. Lastly, the top-laners are largely used to initiate the fights or block the damage incoming from the enemy team. Thus, the top-lane champions have high HP and defensive status, like armor. Tanks are champions capable of receiving high amounts of damage without being killed. Initiators have the responsibility of starting the team combats, while Front-Liners are responsible to stay between the two teams during the combat. Its important to notice that there are heroes that are classified in more than one function, or are suitable in more than one lane. For example, Varus, a LoL champion, can be a mid-laner or ADC, while Malphite can be classified as Tank, Initiator and Mage.

\section{Agent Architecture}

The use of a heterogeneous agent architecture seems reliable for the MOBA domain. Due to the multiple tasks like building, navigating, cooperation and combat, it is possible to use the heterogeneous architecture to implement a multi-tasking agent. Our agent interfaces with LoL through the use of sensors which collect data about the world. The agent also can send action to the hero to perform, modifying the game environment with its actuators, e.g. attacking.

League of Legends has an application programming interface, called Bot of Legends (BoL). BoL retrieves information about the game and performs the same actions as players. This tool is very similar to the Brood War API that is largely used in Starcraft research. Both tools have the same premise: they collect information that is available to players and just perform actions available to players.

The agent is composed of two layers: navigation and micromanagement. Our primary goal is to develop an agent that could navigate and interact in a MOBA domain in a competitive way. That means that we are developing a rational agent with the goal of winning. However, as most agents implemented in the video games domain, our agent is still not capable of outperforming human players.

\begin{figure}
\centering
\includegraphics[width=0.47\textwidth]{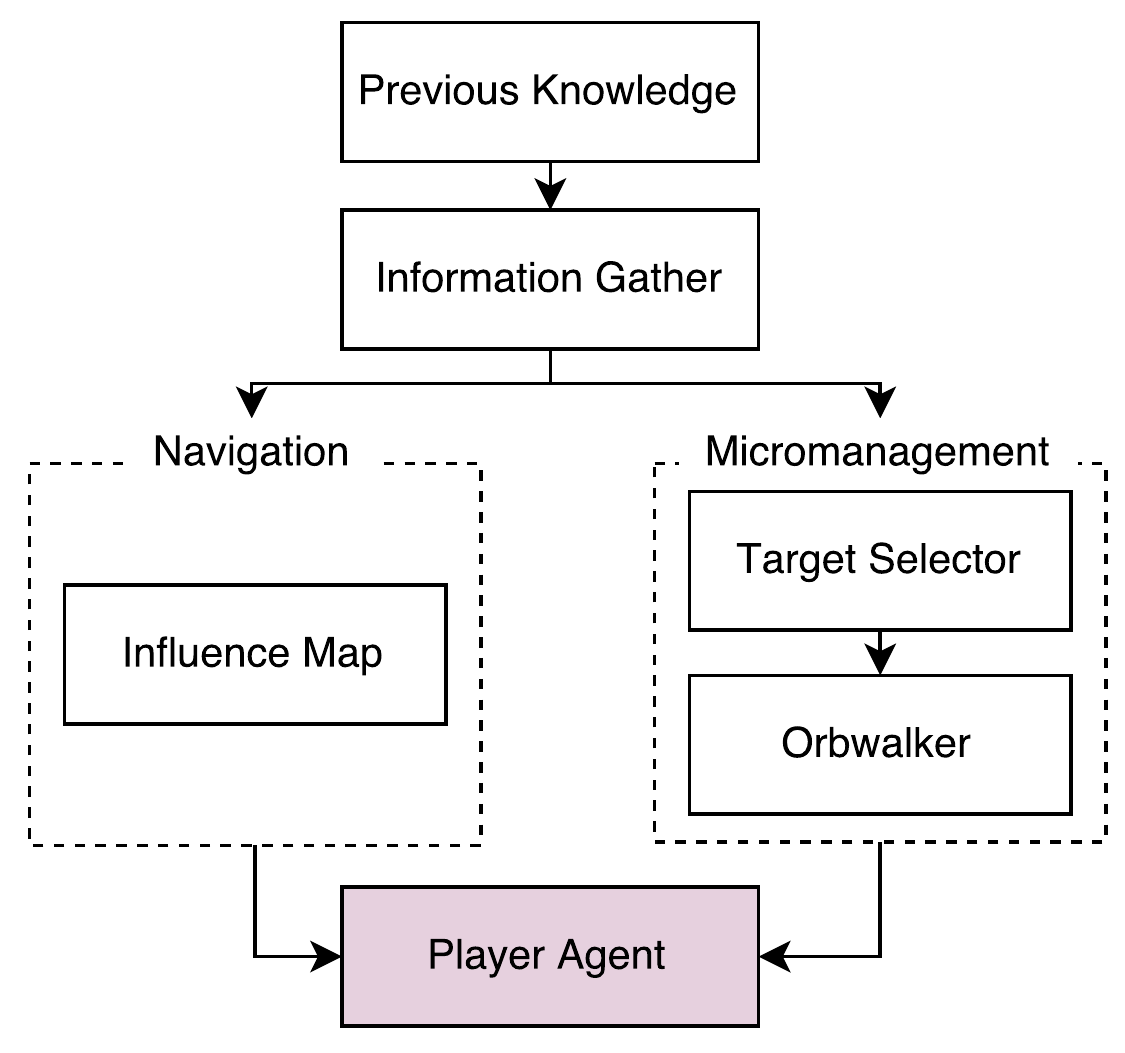}
\caption{Architecture implemented for the agent.}
\label{img:architecture}
\end{figure}

\subsection{Navigation Layer}

The base of our agent is modeling the information collected into valuable information that could be used by all layers. Thus, we inserted a series of hand-authored knowledge and data collected real-time. Further, we model the data collected and hand-authored over an grid-formatted influence map. A visual representation of our model can be seen in Figure \ref{img:architecture}.

The influence map feeds the navigation layer that uses a case-based reasoning (CBR) system. This layer search for best points, taking in account the terrain, dynamic and static features from the domain. The influence maps uses single layer, aiming to improve the agent performance. The IM models knowledge about all agents involved in the MOBA domain. We use mathematical functions to create a single value base in the weight of individuals, generating a 2D grid with decimal values.

Aside from both the IM and the navigation layer, the micromanagement layer uses sensors to perceive the enemy and ally units. Based in the information collected, the micromanagement layer coordinates two CBRs, the target selector and orbwalker. In high-level, target selector is responsible by defining the unit that will be the current target of attacks and abilities. On the other hand, the orbwalker is responsible by reasoning between attack and movement. Both systems will be discussed further.

\subsubsection{Enemy Towers}

When analyzing the towers, in general we observe that it can assume two states: {\em idle} or {\em aggressive}. In the idle state, the tower performs no action, but verifies if there are enemies within its range to be attacked. As soon as an enemy enters the turret range, it will be attacked. In addition, a tower can just attack one enemy at a time, and does not inflict area-of-effect (AoE) damage. 

We can point a preference list for tower enemy choosing order as follows: i) Enemy hero attacking ally hero, ii) any enemy in range, iii) enemy creep and iv) enemy hero. Once the tower selects a target it will not change, unless it dies of comes in first options in the preference list.

In some MOBAs, like Dota or Dota2, the tower prefers the closest enemy, so the players try to stay after their creeps. Using this knowledge, we were able to identify patterns and model an influence level for enemy towers. Different from RTS gameplay, towers in MOBA are very strong damage dealers and have high health values. They are also critical part of the strategy, as they protect the lanes and the structures.

For dealing with enemy towers, we modeled two sub-states for the aggressive state: {\em passive} and {\em active}. The passive state means that there is aggression but the target of the tower is not our hero agent. Whereas the active state means that aggression target is our agent. The influence is then set to the map based in the current state. However in early tests we noticed that our tactical positioning was failing, due the resolution and that our agent was choosing advanced points in the map, an action called {\em tower dive}, causing it to die. We then introduced a weighting system for cells that allowed the agent to choose best positions based on the safest place. Our solution weighted the cells by a $\tau$ variable, that was obtained dividing the enemy unit to agent base distance by the IM cell distance to agent base distance, as seem in the Equation \ref{eq:tau}.

\begin{equation}
\tau = \frac{d_{tb}}{d_{pb}}
\label{eq:tau}
\end{equation}

In addition, when performing tactical analysis, it is important to leave room to risky plays. Sometimes an enemy hero can have a very low hp and the agent could let it go just because it is in tower range. Therefore, when weighting the tower we created a value $\varepsilon$ that allows agent to enter in tower range if necessary. The value $\varepsilon$ should be calculated as the max potential damage to be dealt by the tower. The rules for enemy tower influence can be seen in \ref{eq:enTower}.

\begin{equation}
    w_{p} = \left\{ 
    \begin{array}{l l}
    \left\{ 
        \begin{array}{l l}
            d_{pt}, & \quad d_{pt}<H_{r} - \Delta\\
            \tau * d_{pt},  & \quad H_{r} - \Delta \leq d_{pt} \leq H_{r}\\
            T_{r} - d_{pt}, & \quad \mathrm{else}\\
        \end{array}
    \right., & \quad \alpha \geq 3\\
    \left\{
        \begin{array}{l l}
            -T_{r}, & \quad d_{pt} > \varepsilon \\
            -\infty,  & \quad \mathrm{else}\\
        \end{array}
    \right., & \quad \mathrm{else}
    \end{array}\right.
    \label{eq:enTower}
\end{equation}

\begin{figure}
\centering
\includegraphics[width=0.4\textwidth]{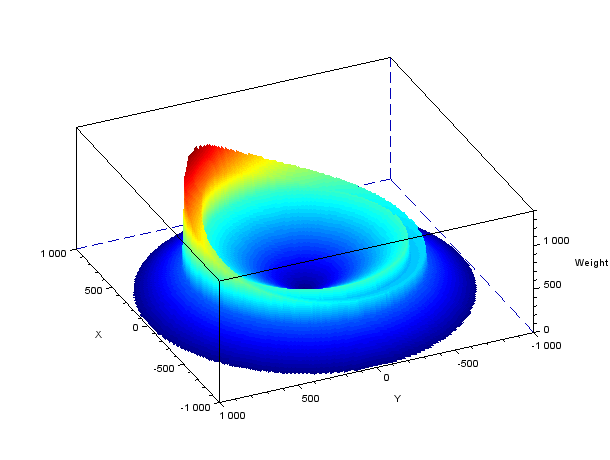}
\caption{Plot of enemy tower in passive aggressive state. Red areas means more desirable positions, while blue areas mean less desirable positions. Note that red area is closer to the base of the agent's team. Base is located in P(1000,1000)}
\label{img:enTower}
\end{figure}

Another problem faced by our agent, and often not discussed in the literature, is the resolution loss problem. As most games dedicates a few computation for AI, it is crucial to have cheap processing in this area \cite{tozour2001}. Therefore, the IM should be tuned aiming to deal a good performance to agent and does not consume a big computational time. The agent could then suffer from low resolution, causing it to perform poorly due to the little amount of data collected by the map. We then introduced a value $\Delta$, that is half of the resolution used, to be used as an error correction. The final plot of equation in passive aggressive state can be seen in figure \ref{img:enTower}. $H_{r}$ means hero range; $d_{pt}$ denotes the euclidean distance between the cell and the tower; $T_{r}$ means the tower range; and $\varepsilon$ denotes a limit area for lowest influence values inside of tower range.

\subsubsection{Enemy Creeps}

Enemy creeps are the main resource collection approach for agents and players. Each time that a player kills an enemy creep the player obtains gold and experience. Thus, there must be a careful analysis of these components, as they can be game changing. The correct positioning around these units will allow the agent to perform the last-hit and, as these units are highly dynamic, they must be observed most constantly.

When calculating the creeps influence we are covering two layers of the agent architecture: movement and micromanagement. In this section, we discuss the movement analysis; the micromanagement will be discussed further, in Section 3.3.1.

\begin{figure}
\centering
\includegraphics[width=0.4\textwidth]{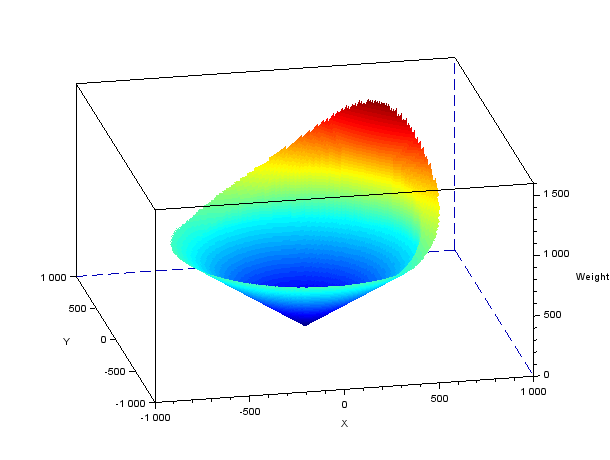}
\caption{Plot of enemy creeps influence. Red areas means more desirable positions, while blue areas mean less desirable positions. Note that the red area is strongly influenced by the $\phi$ variable. Base is located in P(1000,1000)}
\label{img:enCreep}
\end{figure}

To compute the influence of enemy creeps we use the logic proposed by \cite{hagelback2008b}, staying away from the enemy as much as possible but still maintaining it in hero range. As well, as discussed in enemy tower section, we must consider the resolution losses, creating a plateau in our equation, making it possible for the agent to hit the creep. We also consider the weighting relatively to the hero 's own team base, making it stay in the safest place possible, making the agent avoid the damage done by enemy creeps. A plot of the influence done by our model can be seen in Figure \ref{img:enCreep}.

In our early tests we noticed that just by considering this influence with weight was not enough to make the agent perform well in a prerequisite called {\em farming}\footnote{{\em Farming} is the action of collecting resources by executing last-hits in enemy or neutral creeps.}. Making the agent be on the edge of the creeps does not guarantee that the agent will select the creep that will die first, or that have the lowest amount of health points (HP). We then implemented a variable $\phi$, that calculates the creep's missing percent of HP. We inserted $\phi$ in the influence equation that considers the HP of a creep, making the agent to go towards the creep that has the lowest health. Equation \ref{eq:enCreep} shows the influence calculation for enemy creeps. Variable $d_{pm}$ denotes the euclidean distance between the cell and the creep.

\begin{equation}
    w_{p} = 
    \left\{\begin{array}{l l}
        d_{pm}, & \quad d_{pm} < H_{r} - \Delta \\
        max(\tau * (d_{pm}+100 - \phi), w_{p}), & \quad \mathrm{else}
    \end{array} \right.
    \label{eq:enCreep}
\end{equation}

\subsubsection{Ally Towers}

When weighting towers we observe that it is very defensive to be under the range of own's team towers, but for this defense to be effective, the agent should be in a safe area. For example, the agent should be in range of an ally tower and out of range of enemy heroes. However, the safe area is very difficult to abstract due to the diverse mechanics found in MOBA games. Another point that must be observed, is that the player should avoid colliding with towers, as this could lead to time leaks in the navigation. Moreover, in various MOBAs, there are champions that can take advantage of the proximity of the hero to the tower. We then developed an influence equation for allied towers based on these characteristics. The equation provides the tower a linear decay of influence, making its best places near it and the worst ones far away from the tower. For avoiding the collision between the player and the tower, and avoiding to be caught by specific mechanics, we inserted a minimum distance for influence to be set. Thus, the agent will maintain distant from the tower in a secure way, as seen in Equation \ref{eq:allyTower}.

\begin{equation}
    w_{p} = 
    \left\{\begin{array}{l l}
        max(T_{r}-d_{pt}, w_{p}), & \quad d_{pt} > \varepsilon \\
        0, & \quad \mathrm{else}
    \end{array} \right.
    \label{eq:allyTower}
\end{equation}

\subsubsection{Heroes}

Heroes are the most dynamic units in MOBA gameplay, they can move generally faster than creeps and also cast spells and attack. This units are the ones that require more attention when performing tactical analysis in the MOBA domain, due to their high potential of changing the game's tactical landscape. We navigate based strongly in our ally heroes and try to avoid enemy damage.

In our early research phase we implemented the influence range based in the auto-attack range of heroes. That proved to be a bad choice by two main reasons: (a) Melee heroes have a small range, but they can cast spells and deal damage farther and (b) there are heroes that have a smaller range but have a high spell cast range. Therefore, we had to compile a hand-authored knowledge database of all 126 heroes contained in League of Legends. This database informed our agent the distance to be considered as the current hero range. In addition to the hand-authored data, we collected real time information about the spells being cast, allowing the agent to dynamically modify the influence range and perform further analysis about the risk of engaging in a fight against an enemy.

For enemy heroes we consider the danger area as a highly undesirable plateau, so we just attribute the tactical value of that hero, in negative form, in its range area. For ally heroes we perform a very similar approach, applying the calculated tactical value over the cells in range.

\subsection{Combining influences}

Performing multi-agent tactical analysis goes beyond the analysis of each agent separately and running an agent considering it. It is necessary to compile these influences together, making the data collected make sense for the system. The great challenge of making a readable model for the system is to mix them in a way that is computationally efficient.

The initial propose of \cite{tozour2001} is to store various influence map layers and then put them together. Moreover \cite{champandard2011} discusses that if an additional buffer is not used for influence mapping the values can be scattered in a non-desirable way. Thus, in these two approaches we observe that extra memory is needed. When observing the Potential Fields approach proposed by \cite{hagelback2008a} we can observe that more computing resources is needed, as the distance between the cells in the grid must be calculated. 

Our approach uses a similar approach to the proposed by Hagelb{\"a}ck, using the distances to write in a 2D grid. We do not use extra buffering, and, as we do not need any historical information, we do not store these data in our map. All characteristics then drive our approach to a highly tactic map, analyzing just the actual information and reasoning based in that. 

\begin{figure*}[ht]
\centering
\includegraphics[width=1.0\textwidth]{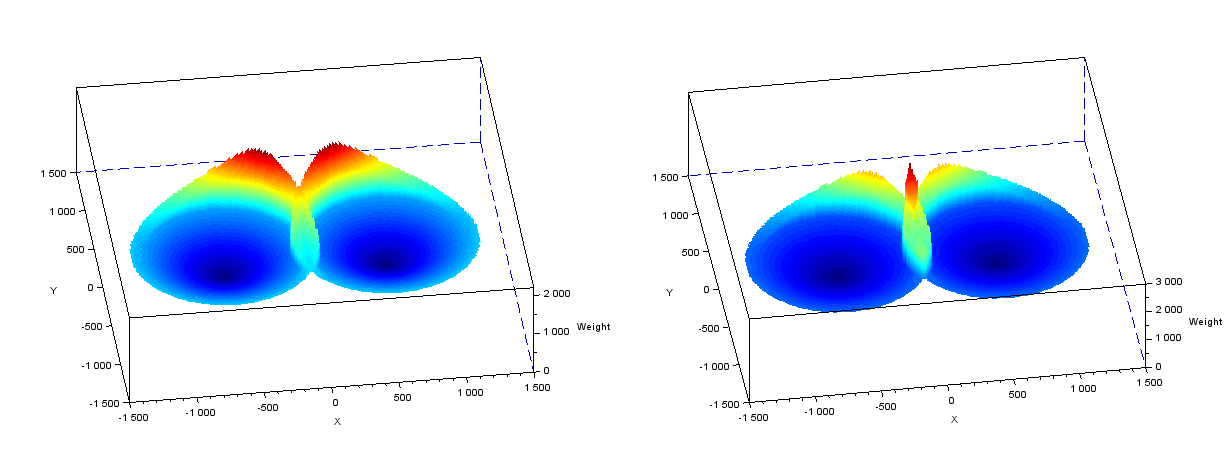}
\caption{Comparing the {\em max}(left) to {\em sum}(right) mixing methods when analysing enemy creeps feature. Red areas means more desirable positions, while blue areas mean less desirable positions. In the right figure we clearly see a local maximum (in red); in the left figure we see a better influence distribution.}
\label{img:enCompare}
\end{figure*}

After all influences are calculated, we need to concatenate them in a single 2D grid. In early IMs, the task of mixing the cells values was done by simple summing each map layer, obtaining rough IMs. However, just summing does not provide a reliable technique for extracting information from this data. For example, summing can drive the agent to a local maximum when analyzing the enemy agent features, as shown in Figure \ref{img:enCompare}. Observe that, while the {\em max} function allows each creep to set its own influence, the sum method tries to sum them together. However, sum creates a local maximum that could expose the agent to dangerous positioning during combat or even to cases where the creeps will be able to defeat the agent due to a local maximum.

To avoid the creation of local maximum, as we are using an Influence Map based in tactical features, we classified the agents involved based in their importance. Doing so would require us to store various Influence Map layers and then perform operations between these layers. However, we figured that we could update each agent class one at a time, obtaining the same results and saving memory. The order was done by the tactical value of each unit. We first update creeps, then we update towers and lastly we update heroes. 

Operation selection was based on the goal of minimizing the local maximum generation and maximizing the outcome of the tactical goal related to that unit. For example, for creeps we aim to maximize the resource collection through last-hitting creeps while we try to minimize the local minimum of creeps, thus the exposition of the agent to danger. Therefore, creeps uses a maximum value function for mixing the values to the map. Our system creates a zero-valued IM in each iteration. Further, our system calculates the influence of creeps, comparing this value to the actual cell value and setting the maximum to current cell.

For the setting of influence from towers, the second feature to be computed, we overwrite the values computed previously. This aims to put the values of towers over the values established by the creeps, showing that it is more important than the previous one.

After setting tower values, we finally calculate and set the values from heroes. As these are the most dynamic unities in the game, they deserve special attention. Even more, defeating a hero can give the agent the necessary time to gain any kind of strategic advantage in the game, so the algorithm prioritizes them.

\subsection{Micromanagement Layer}

Along with a movement layer, it is necessary to implement a layer responsible for coordinating the abilities and attacks from hero. In our approach, as we aim to create a generic technique we do not focus on abilities, but first in simple auto attacks. In the next subsections we discuss the techniques applied in the micromanagement layer.

\subsubsection{Orbwalking and Kiting}

A very useful mechanic in game combat scenarios is to hit and run, especially when the coordinated unit has a higher range than the enemy unit does, this is called kiting. Human players very often perform this kind of mechanics aiming to maximize the damage dealt to enemies both in RTS and in MOBA games. Furthermore, that is an essential technique for damage-based carries and for any agent who aims to farm in MOBA. There is an algorithm, called orbwalker, that is responsible to inform when it is possible to attack or to move, very similar to the approach presented by \cite{uriarte2012}. However, the orbwalking algorithm does not perform any kind of tactical analysis; that is why further scenario analysis is necessary when performing the attack-and-flee mechanic. 

In our approach, we did not implement the kiting algorithm as proposed by \cite{uriarte2012}; however, our agent performed this emerged behavior due to the enemy analysis and positioning system. The orbwalker itself informed the agent when it was possible to attack or when to walk, based in the animation time, turn time and latency of the game. The last parameter showed to be crucial to MOBA approach, as this genre is mostly based in online experience.

As the agent tends to be as far as possible from enemies and maintaining these enemies within the agent range, it is clear that the enemies will be in the edge of enemy attacking field. The agent then receives the repositioning task while attack action is not available, then turns around and attack, when the attack is available. However, our agent does not perform any analysis towards identifying the targets whose kiting can be performed, as the approach presented in \cite{uriarte2012} does.

\subsubsection{Target Selection}

The target selection problem is an open challenge in most strategic games, including RTS and MOBA. The work of \cite{hagelback2008b} address this problem in a very early tactical approach. Later the work of \cite{uriarte2012} discuss this problem in greated detail, assigning tactical values for each enemy based in its distance, manually-assigned tactical threat and DPS. They show that it is an essential part of tactical behavior to select the right enemy to attack. The work of \cite{liu2014} presents an automatic target selection approach using Genetic Algorithms. However, their concept of Target Selection, although still tactic, shows a different concept from \cite{uriarte2012}. Rather than using tactical values or DPS it just considers the health points of each nearby enemy for target selection.

In our approach we collect data and combine them with hand-authored content, building a target selector that analyses the tactical properties of each unit. We use a preference model based in HP, Unit Tactic characteristics and Unit danger potential, also known as {\em aggro}. This mechanism does not have any actuator, it just provides information for the Micromanagement layer, that actuates in the agent state.


As presented, our approach considers the danger first, targeting dangerous enemies, aiming to survive as long as possible. Secondly, it prioritizes resource collection, farming as much as possible. As discussed before, the selection of creeps to be farmed is is a responsibility shared by the micromanagement and movement. Moving next to low HP creeps allows the agent to put them in range, allowing the target selector to aim them in the micromanagement layer and successfully perform the last hit.

Lastly, we verify the existence of a secure scenario to take aim at enemy towers, focusing on goal-controlling and on raising the advantage gained by the agent. By doing so, the agent is aiming to push the lane towards the enemy base, finally defeating the enemy by destroying the main structure.

\section{Empirical Evaluation}

In order to evaluate our approach we performed three  sets of experiments in the domain of League of Legends. The first experiment tests the effectiveness of our agent, testing if the agent is capable of winning in a scenario where it plays alone. Secondly, we test the efficiency of this agent, measuring its performance in terms of resource collection. Lastly we test the agent's performance in a match against human players. These experiments and their results will be discussed in the next subsections.

\subsection{Experiment 1: Winning alone in MOBA}

In this experiment we run the agent in a scenario where it is in a match without ally or enemy heroes. This experiment shows the effectiveness of the agent and the success of the sequence of steps presented to win in a MOBA game. Our baseline is the work of \cite{willich2015}, however we were not capable of replicating the author experiments, as code is not available. In addition, the author uses HoN as game domain, while in this work we use LoL as testbed.

For this experiment we use a map where there is just one lane. In \cite{willich2015} the author uses a similar approach, although they use a three-lane map but considers just one. 

We executed 20 matches with a hero randomly selected from the LoL champion pool. Our agent was capable of winning all matches. In addition, our agent always showed a KDA factor of 0, meaning that it never died and did not present errors. Also for testing the local maximum problem we monitored the agent also collecting the time that the matches lasted. The agent was capable of winning in an average of 22.6 minutes with a standard deviation of 5 minutes. Further, as we were using random heroes, the agent shows consistency for driving various characters.  

We then wanted to measure how well it performed alone in the scenario. A good measure for both the positioning is the resource collection. That leads us to the next experiment.

\subsection{Experiment 2: Resource Collection}

Resource collection in MOBA is different from RTS, as it is performed by killing enemy or neutral creeps, champions and structures. Furthermore, its important to maximize the resource collection. In this test we first run our approach without the $\phi$ variable, that represents the health of creeps. Then we compare the performance of resource collected to the approach with $\phi$. In MOBA community there is a common sense that the best farm score, as performed by professional players, is to last hit 10 creeps per minute, we use that sense as our baseline.

We first ran 10 matches with the agent using the $\phi$ variable for improving positioning and resource collection, then we run 10 matches without using the $\phi$ variable. The tests were executed in the same conditions from Experiment 1, a single lane map without enemy heroes. Our experiments with the health factor shows an average of 92,24\% efficiency in farming (see Table \ref{table:CreepScore}), showing that our agent is capable of efficiently collecting resources. Further, this performance can be compared to the resource collection rate of professional players. In contrast to the previous performance, in matches where the health factor was disabled the efficiency was lowered to an average of 60,84\% of resource collection. 

\begin{table}
    \centering
    \caption{Performance presented by the agent during resource collection}
    \label{table:CreepScore}
    \begin{tabular}{lll}
    \hline
    Method                     & Creeps per Minute & Efficiency(\%) \\ \hline
    Baseline                   & 10                & 100\%          \\
    $\phi$ disabled            & 6.084             & 60.84\%        \\
    $\phi$ enabled             & 9.224             & 92.24\%        \\ \hline
    \end{tabular}
\end{table}

With this experiment we show that our agent is capable of efficiently collecting resources in a highly dynamic environment. It is important to stress that despite the lack of enemies or allies in the match, the agent has to compete with allied creeps for last hits. Moreover, this experiment shows the consistency between the two layers implemented, demonstrating that the combination of navigation and micromanagement system is effective. Further, we show that with the $\phi$ variable we were capable of improve the resource collection task by more than 30\%.

\subsection{Experiment 3: Evaluation Against Human Players}

This experiment deals with a real match in League of Legends. We selected the mode called All Random All Mid (ARAM) to be used as testbed for this experiment. In this mode the map just has one lane and all heroes are selected at random for players that have the chance to exchange them between allies during the hero selection phase. This modes also provides more gold at the beginning of the match and the heroes starts at level three instead of one.

We performed 10 matches in ARAM mode for evaluation and collected the hero performance and win/lose from our agent. We observe that when it plays melee characters the performance is greatly affected, doing much worse than when playing ranged heroes.

The hero shows a high amount of deaths when playing melee champions. When playing ranged champions it shows low death levels, but does not show a high amount of kills. In both melee and ranged, our agent shows a high amount of assists, showing that it helps another players to get kills. In spite of assist characteristics, its not possible to affirm that it behaves in a cooperative fashion, as assists can be granted by just casting an spell or auto attacking. Lastly, as with most agents playing against humans, our agent performed poorly with most heroes. Due the fast dynamics of MOBA, our agent reaction time was not sufficient to play in a team.

\section{Conclusion and Future Work}

In this work we presented an approach for tactical analysis and knowledge modeling in real-time scenario using Influence Maps. For doing so, we used League of Legends, one of the most played games in the world nowadays.

Our experiments results show that this is a promising approach. We implemented the game agent using an heterogeneous architecture composed of two layers: navigation and micromanagement. Furthermore, our approach show that it is possible, and necessary, to strongly connect these layers, as presented by the parameters that connects movement and micromanagement towards resource collection. We also demonstrate that simple features, like a health factor, can greatly increase the performance of the agent. Moreover, further tactical investigation in MOBA domain is required to behave well against humans. Lastly, we obtained a kiting behavior by combining heterogeneous controls. In previous research this behavior required a dedicated module.

A problem that can be tackled in future work is the competition of the current agent against human players. It is known that current state of the arts agents show poor performance when playing against human players, and an interesting question to be answered is whether this happens in MOBA. Moreover, whether the agent performs competitively cooperating with human players, as MOBA requires multiple players to control multiple agents.

Another problem that can be addressed is the development of multiple agents and the interactions between them. Applying approaches of distributing tasks in games, like the presented in \cite{tavares2014}, looks very promising when applied to the MOBA context, as this game genre is strongly oriented to  player roles \cite{yang2014}.

\section{Acknowledgements}

We would like to thank the BoL Community, especially Bilbao and Kenect, that supported us in the development using BoL environment. We also thank the reviewers, that provided insightful feedback and corrections to our paper. This work was supported by CAPES, CNPq and FAPEMIG.

\bibliographystyle{sbgames}
\bibliography{refs}
\end{document}